\documentclass[11pt]{article}
\pdfoutput=1
\usepackage{times}
\usepackage{latexsym}
\usepackage{microtype}
\usepackage{inconsolata}
\usepackage[T1]{fontenc}
\usepackage[utf8]{inputenc}
\usepackage{multirow}
\usepackage{xcolor}

\usepackage[preprint]{coling}

\title{Text Generation Models for Luxembourgish with Limited Data:\\A Balanced Multilingual Strategy}

\author{Alistair Plum \\
  University of Luxembourg \\
  Esch-sur-Alzette, Luxembourg \\
  \texttt{alistair.plum@uni.lu} \\\And
  Tharindu Ranasinghe \\
  Lancaster University \\
  Lancaster, UK \\
  \texttt{t.ranasinghe@aston.ac.uk} \\\AND
  Christoph Purschke \\
  University of Luxembourg \\
  Esch-sur-Alzette, Luxembourg \\
  \texttt{christoph.purschke@uni.lu}\\}

\author{
 \textbf{Alistair Plum$^\diamondsuit$},
 \textbf{Tharindu Ranasinghe$^\spadesuit$},
 \textbf{Christoph Purschke$^\diamondsuit$}
\\
 $^\diamondsuit$University of Luxembourg, Esch-sur-Alzette, Luxembourg \\
 $^\spadesuit$Lancaster University, Lancaster, UK \\
    \{\texttt{alistair.plum,christoph.purschke}\}\texttt{@uni.lu}\\
        \texttt{t.ranasinghe@lancaster.ac.uk}
}

\begin{document}
\maketitle

\begin{abstract}
This paper addresses the challenges in developing language models for less-represented languages, with a focus on Luxembourgish. Despite its active development, Luxembourgish faces a digital data scarcity, exacerbated by Luxembourg's multilingual context. We propose a novel text generation model based on the T5 architecture, combining limited Luxembourgish data with equal amounts, in terms of size and type, of German and French data. We hypothesise that a model trained on Luxembourgish, German, and French will improve the model's cross-lingual transfer learning capabilities and outperform monolingual and large multilingual models. To verify this, the study at hand explores whether multilingual or monolingual training is more beneficial for Luxembourgish language generation. For the evaluation, we introduce \textit{LuxGen}, a text generation benchmark that is the first of its kind for Luxembourgish. 
\end{abstract}

\section{Introduction}
Recent advances in deep learning have made it extremely popular to use language models (LMs)  such as \textsc{BERT} \cite{devlin-etal-2019-bert} and \textsc{T5} \cite{raffel2020exploring} for many tasks in natural language processing (NLP) \cite{LIN2022111}. The tasks range from text classification tasks, such as sentiment analysis \cite{zhang-etal-2024-sentiment} and offensive language detection \cite{OffensEval2023}, to text generation tasks, such as machine translation \cite{zhu-etal-2024-multilingual, wang-etal-2023-document-level} and text summarisation \cite{10.1162/tacl_a_00632, liu-etal-2024-learning}. LMs have achieved state-of-the-art results in many of these tasks \cite{ISLAM2024122666}. 

Even though these LMs are multilingual by design, their support and performance can suffer with languages that are not as well represented \cite{blasi-etal-2022-systematic}. A lot of focus of these LMs tends to fall on English, as well as other high-resource languages \cite{pires-etal-2019-multilingual}. This fact is evidenced by the amount of data used for training certain models. For example, \textsc{mT5} \cite{xue-etal-2021-mt5}, which is trained using the Oscar common crawl data, contains roughly 3.4TB for English, 1.4TB for Chinese, 1.1TB for Russian, 600GB for German and 430GB for Spanish\footnote{\url{https://oscar-project.github.io/documentation/versions/oscar-2301/}}. By contrast, Luxembourgish only has 18MB, and is therefore not well-supported by many current LLMs. However, initial testing has shown that GPT-4o \cite{achiam2023gpt} models can produce Luxembourgish very well, demonstrating the positive effects of cross-lingual transfer-learning \cite{chen-ritter-2021-model} at an immense scale.

More data would certainly improve the situation, but this may not always be possible, meaning that other options need to be considered. Luxembourgish, a West Germanic language, is spoken by around 400,000 people, primarily in Luxembourg \cite{Gilles2019}. In addition to the small number of speakers, Luxembourg is home to a complex societal multilingualism that, historically, has favoured French and German as official languages, especially in formal and written communication. Only since the advent of digital and social media, and as a result of the active language policy to support Luxembourgish, have more significant amounts of text data been produced and therefore made available. As this situation is surely similar with other small varieties, means of finding and using more data are necessary. 

Continuing with the example of Luxembourgish, models such as \textsc{LuxemBERT} \cite{lothritz-etal-2022-luxembert} have used data augmentation, in this case translating from German \cite{olariu-etal-2023-evaluating-data}. Models such as \textsc{LuxGPT2} \cite{bernardy2022} rely on transfer learning. However, none of these methods have provided state-of-the-art performance like models for other languages. In fact, \citet{ranasinghe-etal-2023-publish} has shown that multilingual models outperform existing monolingual Luxembourgish models in a text classification task. We, therefore, propose to fill this gap by combining the largest collection of data for the Luxembourgish language available with carefully considered transfer learning. In this case, in particular, we seek to answer the question of whether similar languages included in the training can improve performance. Recent research has shown that careful selection of multilingual training data improves models for 16 African languages \cite{oladipo-etal-2023-better}. We therefore hypothesise that equal amounts, in terms of size and composition, of Luxembourgish, German, and French will outperform monolingual and multilingual models on Luxembourgish NLP tasks.

In this paper, we address the gap in Luxembourgish NLP and present a novel generative model that is based on the multilingual T5 architecture \cite{xue-etal-2021-mt5}. Moreover, we run multiple evaluations on downstream tasks to ascertain whether multilingual or monolingual pre-training is more beneficial for a Luxembourgish model. To this end, we obtain training data for both German and French, the geographical and socio-cultural neighbours of Luxembourg(ish), and aim to learn more about treating these languages equally in pre-training. In doing so, we also discuss the data compilation, specifically the equivalency of data. For evaluation purposes, due to the current under-represented situation of Luxembourgish NLP, we re-use classification tasks \cite{lothritz-etal-2022-luxembert} and introduce new generative tasks, including news article headline generation, paraphrasing and Wikipedia biography summaries.

Our \textbf{main contributions} are:
\begin{enumerate}
\vspace{-3mm}
    \item Two new generative language models for Luxembourgish, one pre-trained with just Luxembourgish, one pre-trained with Luxembourgish, German and French textual data.\footnote{All material will be available via \url{https://huggingface.co/instilux}}
    \vspace{-3mm}
    \item The introduction of \textit{LuxGen}, the first text generation benchmark for Luxembourgish, including four new text generation tasks.
     \vspace{-3mm}
    \item An evaluation of various language models in terms of performance in Luxembourgish.
     \vspace{-3mm}
    \item Valuable insights into training data composition to increase the performance of a low-resource language.
\end{enumerate}

\section{Related Work}\label{sec:related}
The expansion of language models to encompass European languages beyond English has been a focal point in recent NLP research. The \textsc{T5} \cite{raffel2020exploring}, \textsc{BERT} \cite{devlin-etal-2019-bert}, and \textsc{ELECTRA} \cite{clark2020electra} architectures have been adapted to create not just multilingual \cite{xue-etal-2021-mt5} but also monolingual models that capture the linguistic nuances of individual languages. For example, \citet{chan2020german} developed \textsc{GBERT} and \textsc{GELECTRA}, and later \textsc{GermanT5}, a version of T5 pre-trained exclusively on a large German corpus. Similarly, \citet{le2020flaubert} introduced \textsc{Flaubert}, a French language model based on the BERT architecture. 

In addition to these, \citet{carmo2020ptt5} introduced \textsc{PTT5}, a Portuguese \textsc{T5} model that outperformed previous models in Portuguese text generation and comprehension tasks. \citet{araujo2024sequence} trained a similar \textsc{T5} model for Spanish, and \citet{gudnason-loftsson-2022-pre} evaluated various classification tasks on monolingual Icelandic models based on \textsc{ELECTRA} and \textsc{ConvBERT} \cite{jiang2020convbert}. For the Russian language, \citet{zmitrovich2024familypretrainedtransformerlanguage} developed \textsc{RuT5}, among many others, achieving new benchmarks in Russian NLP tasks. These models underline the importance of tailoring language models to specific languages to capture their unique syntactic and semantic properties.

As a relatively small language, Luxembourgish is less represented in NLP, particularly in comparison to French and German, its socio-cultural neighbours. Luxembourgish has developed in close contact with French and German and today shares grammatical features as well as parts of the lexicon with those languages. This is especially true of German, due to Luxembourgish having developed from the Moselle Franconian dialect \cite{Gilles2019}, but is also true of French. While not typologically, Luxembourgish has been in close contact with French, exhibiting the borrowing of grammatical structures and lexical items, as well as a lot code-switching in written texts. Research in NLP for Luxembourgish has started only recently, with some early exceptions: \citet{adda-decker-etal-2008-developments} introduce various resources for NLP tasks for Luxembourgish; \citet{snoeren-etal-2010-study} analyse typical writing patterns (contextual n-deletion) in written transcripts of speech, and \citet{lavergne-etal-2014-automatic} introduce a manually annotated corpus of mixed language sentences to test a word-based language identification system. Recently, \citet{sirajzade-etal-2020-sentiment} and \citet{Gierschek2022} introduced a state-of-the-art pipeline for sentiment analysis, \citet{purschke2020attitudes} published a pipeline for the automatic orthographic correction of text data and \citet{philippy-etal-2024-forget} introduced a new approach to Zero Shot Classification based on a task-specific dictionary for topic classification. For spoken data, \citet{gilles-etal-2023-asrlux} and \citet{gilles-etal-2023-luxasr} develop \textsc{LUX-ASR}, a performant Automatic Speech Recognition for Luxembourgish. For automatic comment moderation \cite{ranasinghe-etal-2023-publish} and orthographic normalization \cite{lutgen2024}, models based on the T5 architecture have shown to perform well for Luxembourgish.

Looking at language models for Luxembourgish, various strategies have been tested. Some have been trained using transfer-learning from German, such as \textsc{LuxGPT} \cite{bernardy2022}. Another model for Luxembourgish exists in \textsc{LuxemBERT} \cite{lothritz-etal-2022-luxembert}, which used augmentation techniques to produce more Luxembourgish data, and which performs on par with multilingual BERT on Luxembourgish language tasks such as part-of-speech tagging, named entity recognition and news classification \cite{lothritz2023comparing}. To this end, the authors of the afore mentioned model also introduced the corresponding resources. \citet{anastasiou-2022-enrich4all} introduced \textsc{ENRICH4ALL}, another BERT model for the development of a multilingual chatbot in administrative contexts, trained with a specifically annotated corpus. These models all demonstrate the various strategies used to create the models. 

Nevertheless, it is clear that not enough resources for Luxembourgish exist yet, and that research related to NLP in Luxembourgish is limited. Both of these reasons explain why there are only some benchmarks for classification tasks in Luxembourgish, and none for generative tasks. The absence of benchmarks, small amounts of data, and lack of more generative models are areas we hope to address with this research.

\section{Data}\label{sec:data}
Data is essential to training a language model and plays a critical role in this research, as outlined in the introduction. Because we wanted to investigate specifically the composition of the data used for training, we explain the choices made in this section.

It would be reasonable to assume a typical approach of simply using all the data one can find. Because Luxembourgish only has a minimal amount of data available compared to German and French, this would lead to a large imbalance in the data, calling into question whether the Luxembourgish part would even be worth including. As stated previously, we therefore aimed to match the German and French data largely to mirror that of the Luxembourgish data in terms of size, type, and domain. We consciously set out to collect roughly equal amounts of data for each language to test specifically how much better the language model would perform in comparison to a monolingual Luxembourgish model. Table \ref{tab:token_counts} presents an overview of the different areas and the number of tokens from these areas, which will be described in the following paragraphs for each language. 

\begin{table}[h!]
    \centering
    \begin{tabular}{|l||c|c|c|}
    \hline
    \textbf{Domain} & \textbf{LB} & \textbf{DE} & \textbf{FR} \\ \hline\hline
    Radio & 17,5M & 1,3M & - \\ \hline
    News & 42,5M & 71,7M & 62,1M \\ \hline
    Parl. & 17,4M & 31,2M & 40,0M \\ \hline
    Web & 84,9M & 56,2M & 83,0M \\ \hline
    Wiki & 7,4M & 16,4M & 18,6M \\ \hline\hline
    \textbf{Total} & \textbf{169,7M} & \textbf{176,8M} & \textbf{203,7M} \\ \hline
    \end{tabular}
    \caption{Token counts for texts from each domain for each language covered in the training data selection.}
    \label{tab:token_counts}
\end{table}

\paragraph{Luxembourgish}
For Luxembourgish, we opted to compile the dataset ourselves, as opposed to just using crawl data. This is due to reasons pointed out in previous sections, as well as affording us more control over the incoming data and allowing a more controlled compilation of corresponding data in the other languages. To this end, we aimed to collect all data as it is available to us (un-normalized) from known collections. This includes news articles (News), transcribed radio interviews (Radio) and user comments (Coms.) from the country's largest public news broadcaster, RTL Lëtzebuerg (RTL). As seen in Table \ref{tab:token_counts_lb}, this forms the bulk of the Luxembourgish data in terms of the number of tokens. The RTL data spans the years 2008 until 2023 and has been used for many other Luxembourgish NLP research \cite{lothritz-etal-2022-luxembert,lothritz2023comparing,Gierschek2022,purschke2020attitudes,ranasinghe-etal-2023-publish}. In terms of news related text, we also added data from the Leipzig (Lpzg) collection \cite{goldhahn2012building}
which includes data from other Luxembourgish news sites.\footnote{\url{https://corpora.uni-leipzig.de/}}

For web content, we also made use of data from the Leipzig collection
using specifically 1 million sentences from the latest web crawl, excluding RTL.
In addition, we use text from Luxembourgish chat rooms. Encyclopaedic text was also used in the form of Wikipedia articles, which we obtained from the latest dump as of the time of training. This spans roughly 70,000 articles about various topics, but mainly biographies and information about locations. Similarly, we used all example sentences from the Luxembourgish online dictionary (LOD).\footnote{\url{https://lod.lu}} Finally, political speeches and debates are also represented in the corpus, which are transcribed for the Chambre des Députés (Chamber), the national legislature chamber of Luxembourg.

\begin{table}[h!]
    \centering
    \begin{tabular}{|l||c|c|c|}
    \hline
    \textbf{Resource} & \textbf{Tokens} & \textbf{Types} & \textbf{TTR} \\ \hline\hline
    RTL Radio & 17,5M & 741,000 & .0423 \\ \hline
    RTL News & 36,7M & 1,46M & .0398 \\ \hline
    RTL Coms. & 55,8M & 2,58M & .0463 \\ \hline
    Lpzg. News & 5,85M & 677,000 & .1158 \\ \hline
    Lpzg. Web & 17,1M & 1,83M & .1074 \\ \hline
    Chat Logs & 12,1M & 659,000 & .0545 \\ \hline
    Chamber & 17,4M & 404,000 & .0233 \\ \hline
    Wikipedia & 6,87M & 576,000 & .0839 \\ \hline
    LOD & 0,5M & 44,000 & .0874 \\ \hline\hline
    \textbf{Total} & \textbf{169,73M} & \textbf{8,87M} & \textbf{.0628} \\ \hline
    \end{tabular}
    \caption{Token counts, type counts, and Type-Token Ratio (TTR) for Luxembourgish language resources.}
    \label{tab:token_counts_lb}
\end{table}

\paragraph{German}
For German, we aimed to collect the closest corpora to the Luxembourgish corpora. Starting with news related text, we again made use of parts of the Leipzig corpus, specifically from German news sites (in roughly equal token quantities) as well as the Ten Thousand German News Articles Dataset (10kGNAD) dataset\footnote{\url{https://tblock.github.io/10kGNAD/}}, which consists of over 10,000 articles from an Austrian newspaper across nine topics. As such, these articles are the unused part of the One Million Posts Corpus (OMPC) \cite{Schabus2017}, which we use to replicate the situation in Luxembourgish, where we have news articles and the user discussion under each article. We also use the Potsdam Commentary Corpus (PCC) \cite{bourgonje2020potsdam} to add further user comments from German newspaper websites. For the transcribed radio interviews in Luxembourgish, we found the closest equivalent in the German Radio Interviews (GRAIN) corpus \cite{schweitzer2018german}, which comprises a small amount of transcribed radio interviews.

For web content, we used the Leipzig Corpus web crawl data for German. We could not replicate chat room data for German, and decided to leave this, as it only makes up a small amount of the Luxembourgish data. For encyclopaedic text we naturally used Wikipedia again, selecting the Leipzig Corpus Wikipedia selection for ease of use, and since one year is almost equivalent to the whole Luxembourgish Wikipedia corpus. To add political speeches and debates, we used the German section of the Digital Corpus of the European Parliament (DCEP) \cite{hajlaoui2014dcep}, specifically the AGENDA, IM-PRESS, MOTION and REPORT subsections, as these contained the most relevant textual data.

\paragraph{French}
For French, we also aimed to collect the most related textual data; however, we found the situation to be somewhat different to German, with not as many resources easily available. We used again Leipzig Corpora for news, French News 2010 and 2022 1M sentences, as well as French Newscrawl 2020 1M sentences. In addition, we used French Mixed Typical 2012 1M sentences to represent typical web data. To supplement web data and more comment style content, we used the French Reddit dataset from Kaggle.\footnote{\url{https://www.kaggle.com/datasets/breandan/french-reddit-discussion}}

As for both previous languages we used an extract of Wikipedia articles for encyclopaedic text, making use of the Leipzig Corpora yet again, the French Wikipedia 2021 1M sentences collection. For political speeches and discussion, we used the French section of the DCEP, with the same subsections as for German.

\begin{table*}[!ht]
\centering
\begin{tabular}{|l||r|r|l|l|}
\hline
\textbf{Task} & \textbf{Train} & \textbf{Test} & \textbf{Type} \\
\hline \hline
News Title & 162,882 & 13,852 & RTL news articles \\ \hline
Positive comment & 3,236 & 810 & RTL articles \& comments \\ \hline
Negative comment & 3,236 & 810 & RTL articles \& comments \\ \hline
Description & 11,858 & 2,094 & Wikipedia articles \\
\hline
\end{tabular}
\caption{Overview of the data for the four different LuxGen tasks, including no. of training and test instances, as well as the types of instances.}
\label{tab:downstream_overview}
\end{table*}

\section{Models}\label{sec:models}
We leverage our unlabelled data described in Section \ref{sec:data} to pretrain two models: \textsc{LuxT5} on Luxembourgish data and \textsc{LuxT5-Grande} on Luxembourgish, German and French data using the \textsc{T5-Base} encoder-decoder architecture \cite{raffel2020exploring}. Each of the encoder and decoder components contains 12 layers, each with 12 attention heads and 768 hidden units. In total, this results in a model with 220 million parameters. 

We used the same objective as the original \textsc{T5} models \cite{raffel2020exploring}. The main idea is to feed the model with corrupted (masked) versions of the original sentence and train it to reconstruct the original sequence. This denoising objective works by randomly sampling and dropping out 15\% of tokens in the input sequence. All consecutive spans of dropped-out tokens are then replaced by a single sentinel token.

For both of our pre-trained models, we use a learning rate of 1e-4, a batch size of 128 sequences, and a maximum sequence length of 512. We pre-train each model for 1M steps.

\section{\emph{LuxGen}: Text Generation Benchmark for Luxembourgish}\label{sec:eval}
To evaluate the \textsc{LuxT5} and \textsc{LuxT5-Grande} models, we defined \emph{LuxGen}: A text generation benchmark for Luxembourgish, consisting of four text generative tasks. Due to data being limited for Luxembourgish, especially in terms of benchmarks, we derive these tasks mainly from RTL data, as this already has the most metadata available. The generative tasks are all novel for Luxembourgish. We believe these tasks to be the best currently available to evaluate the performance of text generation models, including recent large language models for NLP tasks in Luxembourgish. An overview of the available data is presented in Table \ref{tab:downstream_overview}.

\subsection{News Headline Generation}
In this task, the model is trained to generate a headline for a specific news article. We created this task as it offered itself as a straight-forward task from the data. Similar tasks have been proposed by \citet{hettiarachchi-etal-2024-nsina-news}, \citet{nagoudi-etal-2022-arat5}, and \citet{aralikatte-etal-2023-varta}.

We used news articles taken from the RTL collection that we used for pre-training the models. It should be made clear at this point that we removed the article headlines at the point of pre-training so that we could obtain unbiased results. The exact number of training and testing instances can be seen in Table \ref{tab:downstream_overview}.

\subsection{Positive and Negative Comment Generation}
For this task, we utilise user voting on the RTL user comments dataset to extract the most upvoted and downvoted comments. Using the corresponding RTL article that a given user comment was made on, the task is to generate the most upvoted and the most downvoted comment.

The datasets used for this task are the RTL user comments dataset and the RTL news articles dataset. Matching the comments with the corresponding article by ID, we then calculate the up/down ratio and determine the most upvoted and most downvoted comments. Since the voting on user comments feature was only introduced in 2019, our data is partially limited for this task, especially as not every article has user comments that have votes. The comments have also been moderated by RTL to remove harmful or offensive language and anonymise users. We have 4044 comments each for most upvoted and most downvoted, utilising an 85\% to 15\% train test split in order to retain the maximum amount for training (see also Table \ref{tab:downstream_overview}).

\begin{table*}[!ht]
\centering
\begin{tabular}{|l|l||c|c|c|c|}
\hline
\textbf{Group} & \textbf{Model} & \textbf{Headline} & \textbf{Positive} & \textbf{Negative} & \textbf{Wiki} \\ \hline \hline
\multirow{2}{*}{Prompt} & \textsc{GPT-4o-2024-05-13} & 0.0482 & 0.0032 & 0.0017 & 0.1001 \\
 & \textsc{Llama-3.1-8B-Ins.} & 0.0359 & 0.0037 & 0.0028 & 0.0268 \\ \hline
\multirow{2}{*}{Pre + Fine} & \textsc{LuxT5-Grande} & \textbf{0.2130} & \textbf{0.0810} & \textbf{0.0780} & \textbf{0.1100} \\ 
 & \textsc{LuxT5} & 0.1680 & 0.0450 & 0.0320 & 0.0280  \\ \hline
\multirow{4}{*}{Fine-tuning} & \textsc{mT5-base} & 0.1820 & 0.0009 & 0.0006 & 0.0230  \\ 
 & \textsc{mT5-small}  & 0.1650 & 0.0003 & 0.0003 & 0.0160   \\ 
 & \textsc{ByT5-base}  & 0.0310 & 0.0000 & 0.0000 & 0.0002 \\ 
 & \textsc{ByT5-small}   & 0.0320 & 0.0000 & 0.0000 & 0.0001  \\  \hline
\end{tabular}
\caption{BLEU scores for different tasks in \emph{LuxGen}. The best result for each task is in bold.}
\label{tab:results}
\end{table*}

\subsection{Short Description Generation} 
We define the final generative task as description generation. We utilise Wikipedia and its structured equivalent, Wikidata, for this task. The task for the model is to generate a short description of a Wikipedia article. 

For this task, we use all Luxembourgish Wikipedia articles that have a short description on Wikidata. These descriptions can almost be seen as short labels; nevertheless, we use this data for a generative task. As the number of articles in Luxembourgish is quite small, we collected roughly 14,000 articles with descriptions. An exact overview of the training and test instances is presented in Table \ref{tab:downstream_overview}.

\section{Evaluation}
In this section, we evaluate the performance of our proposed models, \textsc{LuxT5} and \textsc{LuxT5-Grande}, on Luxembourgish text generation tasks encompassed in the \emph{LuxGen} dataset, and a classification task for Luxembourgish introduced by \citet{ranasinghe-etal-2023-publish}. We compare our models against several baselines, including the non fine-tuned large language models (LLMs) \textsc{Llama 3} \cite{dubey2024llama}, \textsc{GPT-4o} \cite{achiam2023gpt}, and \textsc{Mistral}, as well as fine-tuned versions of mT5 \cite{xue-etal-2021-mt5} and \textsc{ByT5} \cite{xue-etal-2022-byt5}. Our evaluation comprises both automatic metrics, using BLEU scores \cite{10.3115/1073083.1073135} due to the lack of advanced NLG metrics for Luxembourgish, standard metrics for classification, and a manual analysis to provide a comprehensive understanding of each model's capabilities in generating accurate and fluent Luxembourgish text.

\subsection{LuxGen}
For all tasks in \emph{LuxGen}, we compare \textsc{LuxT5} and \textsc{LuxT5-Grande} to Llama 3 and GPT 4o (non fine-tuned), as well as several fine-tuned variants of mT5 \cite{xue-etal-2021-mt5} and ByT5 \cite{xue-etal-2022-byt5}. All the tasks were considered sequence-to-sequence tasks. For all the T5-based models, we used the same configurations: a batch size of 8, Adam optimiser with learning rate 1e-4, and a linear learning rate warm-up over 10\% of the training data and trained the models over ten epochs. For Llama 3 and GPT 4o, we used prompts in English and optimised them to achieve the best output. Exact prompts are listed in Appendix \ref{sec:appendix}. \textsc{Mistral} did not produce any outputs in Luxembourgish.

For the automatic evaluation, we utilised BLEU score \cite{10.1145/3305260}. While there are advanced NLG metrics such as BLEURT \cite{sellam-etal-2020-bleurt} and BERTScore \cite{zhang2019bertscore}, they do not currently support Luxembourgish. The results are shown in Table \ref{tab:results}. As can be seen in Table \ref{tab:results}, \textsc{LuxT5-Grande} outperforms LuxT5 and other baselines in all the tasks in \emph{LuxGen}. The key findings of the results are listed below. 

\paragraph{\textsc{LuxT5-Grande}} As stated previously, \textsc{LuxT5-Grande} outperforms all models in the \emph{LuxGen} tasks. For the tasks with more training instances, such as headline generation, the gap between the mT5 models and \textsc{LuxT5-Grande} is low. However, for the tasks where the number of training instances is lower, there is a larger gap between \textsc{LuxT5-Grande} and the mT5 models. We believe that when there are a large number of training instances, \textsc{mT5} can come close to specific T5 models. However, they are unable to train their weights properly when there are fewer training instances. We also see that the LLMs do not perform as well, except \textsc{GPT-4o} in the Wikipedia description task, which could very well be due to the overlap of training data, i.e. GPT having seen Wikipedia in training.

\begin{table*}[!ht]
\centering
\begin{tabular}{|l|ccc|ccc|ccc|c|}
\hline
\multirow{2}{*}{\textbf{Model}} & \multicolumn{3}{c|}{\textbf{Archived}} & \multicolumn{3}{c|}{\textbf{Published}} & \multicolumn{3}{c|}{\textbf{Weighted Average}} & \multirow{2}{*}{\textbf{F1 Macro}} \\
& \textbf{P} & \textbf{R} & \textbf{F1} & \textbf{P} & \textbf{R} & \textbf{F1} & \textbf{P} & \textbf{R} & \textbf{F1} &  \\ \hline \hline
\textsc{mBERT} & 0.58 & 0.06 & 0.12 & 0.77 & 0.97 & 0.86 & 0.72 & 0.77 & 0.70 & 0.49 \\
\textsc{LuxemBERT} & 0.60 & 0.08 & 0.15 & 0.78 & \textbf{0.98} & 0.87 & 0.73 & 0.77 & 0.70 & 0.51 \\
\textsc{ByT5 Large} & 0.67 & \textbf{0.20} & \textbf{0.31} & \textbf{0.79} & \textbf{0.98} & \textbf{0.88} & \textbf{0.77} & 0.78 & \textbf{0.74} & \textbf{0.59} \\
\textbf{\textsc{LuxT5-Grande}} & \textbf{0.69} & 0.15 & 0.25 & \textbf{0.79} & \textbf{0.98} & \textbf{0.88} & \textbf{0.77} & \textbf{0.79} & 0.73 & 0.56 \\
\hline
\end{tabular}
\caption{Results for the moderation classification task.}
\label{tab:classification_results}
\end{table*}

\paragraph{Monolingual Training} \textsc{LuxT5}, which we only trained using Luxembourgish data, does not consistently outperform mT5 models in \emph{LuxGen}. We believe that this shows the Luxembourgish data on its own is simply insufficient. Since there was not much data to train \textsc{LuxT5}, the model might be inconsistent in some tasks. This is also shown by the LLM results, which do not reach the performance of \textsc{LuxT5}, but come close even without fine-tuning (but having many times more data in pre-training).

\paragraph{\textsc{ByT5} models} Previous research suggested that ByT5 models will perform well in Luxembourgish tasks \cite{ranasinghe-etal-2023-publish}. Surprisingly, the results in Table \ref{tab:results} suggest that ByT5 models perform poorly in Luxembourgish text generation tasks. We assume that this is due to the model architecture not being as well suited to generative tasks in the \emph{LuxGen} settings.

\subsection{Classification Evaluation}
The classification task we evaluate involves predicting whether a given user comment is \textit{Archived} or \textit{Published}, as first introduced by \citet{ranasinghe-etal-2023-publish}, for which we reproduced the same data splits for comparability. This task presents significant challenges due to class imbalance and the subtle distinctions between the two categories, making it a valuable benchmark for assessing model performance on Luxembourgish text classification.

The results presented in Table \ref{tab:classification_results} indicate that our proposed model, \textsc{LuxT5-Grande}, outperforms the previously released Luxembourgish model\textsc{LuxemBERT} \cite{lothritz-etal-2022-luxembert} across multiple evaluation metrics. Specifically, \textsc{LuxT5-Grande} achieves higher precision for the "Archived" class and a better weighted average precision. While \textsc{ByT5 Large} attains the highest overall performance with an F1 Macro score of 0.59, \textsc{LuxT5-Grande} comes close with an F1 Macro of 0.56, despite not being optimized for classification tasks.

We attribute the superior performance of \textsc{ByT5 Large} in this classification task to its architecture, which is well-suited for handling character-level information — a critical factor for languages with rich morphological structures like Luxembourgish. The byte-level tokenization employed by \textsc{ByT5 Large} enables it to capture subtle textual nuances essential for accurate classification. In contrast, \textsc{LuxT5-Grande} was primarily developed as a generative model. Despite this, its competitive performance underscores the effectiveness of our pre-training and fine-tuning approach. With further optimization targeted at classification, \textsc{LuxT5-Grande} has the potential to surpass existing models by combining strong generative capabilities with robust classification performance.

\subsection{Manual Evaluation}
Due to the fact that the BLEU score does not offer complete insight into the performance of the models for Luxembourgish, we also completed a manual investigation and evaluation of some sample sentences to gain more insight into the generated text. As we observed during the evaluation, the two evaluated LLMs produced seemingly good Luxembourgish, although often having sentences with reversed logic to that of the real test output (for example, \textit{team B lost to team A} versus \textit{team A won against team B}). Because of BLEU's evaluation on word alignment, this impacts the scores heavily. Therefore, we have opted to include our analysis of a sample of the output predictions for \emph{LuxGen}. Since our pre-trained \textsc{LuxT5-Grande} model performed best in terms of BLEU out of the T5-based models, we selected its output, alongside the monolingual \textsc{LuxT5} to directly compare the effects of adding more languages in pre-training, as well as the two LLMs, for further analysis.

For the analysis, we took 20 random sentences per task in \emph{LuxGen}, and evaluated the predictions by taking three categories into account: \textit{task}, \textit{content}, and \textit{correctness}. With \textit{task}, we checked whether the task had been completed, e.g. \textit{has a headline been generated? Is the length appropriate?}. For \textit{content}, we compared not only the target text but also the input text to check whether the model has reproduced appropriate content or not. Finally, for \textit{correctness}, we checked the output according to Luxembourgish grammar and orthography rules, as well as whether the model stayed in the correct language. We summarise our findings per model group.

\paragraph{LLMs} As part of the evaluation, we prompted \textsc{GPT-4o} and \textsc{Llama-3.1-8B-Instruct} to generate predictions for \emph{LuxGen}. Looking at the outputs, we consistently saw that both models were able to complete the assigned tasks, which is to be expected. It was observed that although both models tended to generate texts that were longer than the target outputs (see example outputs below), the tasks were still completed, usually displaying more information than the targets. This relates also to content, which was generally reproduced correctly, although headlines tended to contain much more information than the target headlines, but the articles did contain this information. In terms of correctness, GPT-4o was mainly able to produce correct Luxembourgish, only rarely switching to German or hallucinating Luxembourgish forms, as indicated in the example outputs in red. \textsc{Llama-3.1-8B-Instruct}, on the other hand, did not produce correct Luxembourgish, often switching mid-way through predictions to German, highlighted in the example sentence below in blue. Compared with the other two models, the LLMs produced more passive constructions than active ones. On the whole, it is fair to say the BLEU scores do not accurately reflect the quality of the output of these models in terms of task and language.

\textbf{Example Outputs:}
\begin{itemize}
    \item \texttt{Llama:} Bolivien: Dausende \textcolor{blue}{Polizisten jagen international gesuchten} Drogeboss Sebastian Cabrera.
    \item \texttt{GPT:} Dausende Polizisten an Bolivien op déi \textcolor{red}{groﬂ Botter} - International gesichte Drogeboss entkommt nach ëmmer.
    \item \texttt{LuxT5:} Dausende Poliziste \textcolor{olive}{kämpfe géint} Drogeboss Sebastian Cabrera.
    \item \texttt{Original:} Police \textcolor{olive}{sicht no} Drogeboss Sebastian Cabrera.
\end{itemize}

\paragraph{LuxT5s} Looking at the results of our \textsc{LuxT5} models, we saw that both models were able to complete the various tasks. In fact, both models generated outputs that were much more similar in length and style to the target outputs, which is to be expected due to these models being fine-tuned for the various tasks. In terms of content, we often found that \textsc{LuxT5} would often add random bits of information that it did not reproduce from the text inputs, making it factually incorrect in places. \textsc{LuxT5-Grande} did not suffer from this, demonstrating that adding more language data in addition to the Luxembourgish base to be beneficial. We also saw that both models do not switch around the sentence logic, as observed with the LLMs, but did slightly change the meaning, as highlighted in the example outputs above in gold. Finally, in terms of correctness, both models generated Luxembourgish without switching to German, with only minor mistakes. It should be noted, however, that both models suffered slightly from finishing the generation too early, therefore leaving unfinished words in places.

\paragraph{Overall} We saw that all four models produced much better output than the automatic evaluation would indicate. It seems clear that this has much to do with the fact that the outputs, while addressing the task, being factually correct, and linguistically correct, often look nothing like the target predictions. Because this would mean that many words are misaligned, the BLEU scores would suffer from this. Although this is not optimal, the targets that we have are the only ones that we can work with that have been produced by real humans. Nevertheless, with these results, we see that including similar languages in the pre-training process can improve the performance in language models.

\section{Conclusion}\label{sec:conclusion}
This paper has demonstrated the performance of multiple language models for the Luxembourgish language. The models demonstrated are all capable of both text classification and text generation tasks. While we have presented a detailed evaluation of the various models, we have also described the different datasets with which the models were trained. In doing so, we have shown that large, massively multilingual models do not necessarily perform best for small, low-resource languages. In fact, we think it is clear that smaller models, trained on the limited amount of data available for a given low-resource language, can benefit from the addition of equal amounts of linguistically related languages. Our results indicate that such models can outperform larger multilingual language models consistently and can come very close to the performance of LLMs, like GPT, although at a considerably lower cost in terms of training data size and training time. With these findings in mind, we plan ablation studies in the future to determine the exact effects more precisely.

The findings of this paper further suggest that there may be positive implications for not just low-resource languages that can benefit from socio-cultural neighbours or contact languages, but also for all kinds of varieties within a given language, such as regional dialects. Nonetheless, these findings require further research, which will shape our future outlook on the topic of this paper. We plan to experiment with adding and removing further languages to and from our models to assess the performance impact. We also want to look more closely at the precise quantities and composition of added data, as well as the balances in relation to other languages. Furthermore, we plan to test our approach for regional varieties of German and other languages, to determine whether this approach of adding linguistically related languages or language varieties to a model can help performance.

\section*{Limitations}
It is clear that using BLEU score is not an ideal metric, especially given the fact that Luxembourgish is not widely standardised in practice, meaning that character variation is always present, making a character-based metric difficult to interpret for evaluation. However, due to Luxembourgish being a low-research language, we could not determine a more suitable metric. The fact that there is limited data for Luxembourgish, and that there is only a tiny amount of human annotated data, exacerbates this problem. 

\section*{Ethics Statement}
This research was conducted using existing annotated datasets and did not involve the creation of any new human annotations. All data utilized in this study was previously publicly available and did not require any new data collection. The datasets employed in this research are properly licensed for this use.

\bibliography{custom}

\appendix
\section{List of Prompts}
\label{sec:appendix}

\subsection{GPT-4o}

\begin{itemize}
    \item You are an editorial assistant for a Luxembourgish news outlet. Your task is to generate a news headline for a news article, based on the content of the article.
    \item You are a Luxembourgish social media user. Your task is to generate a positive user comment in response to a news article. The comment should be closest to a comment that is most likely to get the most upvotes or thumbs up from other users.
    \item You are a Luxembourgish social media user. Your task is to generate a user comment in response to a news article. The comment should be closest to a comment that is most likely to get the most downvotes or thumbs down from other users.
    \item Based on a Luxembourgish Wikipedia article as input, your task is to generate a short description in Luxembourgish of the thing that is being described. The description can be as short as a word, and no longer than a short sentence.
\end{itemize}

\subsection{Llama 3}

\begin{itemize}
    \item You are an editorial assistant for a Luxembourgish news outlet. Your task is to generate a news headline for the following news article, based on the content of the article. Only return the title.
    \item You are a Luxembourgish social media user. Your task is to generate a user comment in response to a news article. The comment should be closest to a comment that is most likely to get the most upvotes or thumbs up from other users. Only return the comment.
    \item You are a Luxembourgish social media user. Your task is to generate a user comment in response to a news article. The comment should be closest to a comment that is most likely to get the most downvotes or thumbs down from other users. Only return the comment.
    \item Based on a Luxembourgish Wikipedia article as input, your task is to generate the corresponding short Wikipedia description in Luxembourgish of the thing that is being described. The general description should not be longer than a couple of words. Only return the description.
\end{itemize}

\end{document}